\documentclass[fleqn]{article}
\pdfoutput=1
\usepackage{url}            
\usepackage[T1]{fontenc}    
\usepackage[utf8]{inputenc} 
\usepackage[english]{babel} 
\usepackage{csquotes}
\usepackage[]{natbib}
\usepackage{amsmath}        
\usepackage{mathrsfs}       
\usepackage{amssymb}        
\usepackage{amsfonts}       
\usepackage[svgnames]{xcolor} 
\usepackage{geometry}       
\usepackage{enumitem}
\usepackage{graphicx}
\usepackage{gb4e}
\noautomath
\usepackage{CJKutf8}
\usepackage{algorithm}
\usepackage{algorithmic}
\usepackage{qtree}
\usepackage{comment}
\usepackage{tikz} 
\usepackage{tikz-dependency}
\usetikzlibrary{patterns,arrows,decorations.pathreplacing}

\title{Neural machine translation, corpus and frugality}
\author{Raoul Blin, CNRS-CRLAO, blin@ehess.fr}

\begin{document}

\maketitle

\begin{abstract}
In machine translation field, in both academia and industry, there is a growing interest in increasingly powerful systems, using corpora of several hundred million to several billion examples. 
These systems represent the state-of-the-art. 
Here we defend the idea of developing in parallel \enquote{frugal} bilingual translation systems, trained with relatively small corpora. 
Based on the observation of a standard human professional translator, we estimate that the corpora should be composed at maximum of a monolingual sub-corpus of 75 million examples for the source language, a second monolingual sub-corpus of 6 million examples for the target language, and an aligned bilingual sub-corpus of 6 million bi-examples. A less desirable alternative would be an aligned bilingual corpus of 47.5 million bi-examples. 
\end{abstract}

\section{Big is beautiful ?\label{intro}}

In the \enquote{real world} (versus the experimental world), Deepl is reputed to be one of the best automatic translators as far as European languages are concerned, if not the best. 
Internal measurements \footnote{The company announces a BLEU score of 33.1 on the English-German WMT 2014 test; https://www.deepl.com/press.html, 2020/06/17. Note that a preliminar version of the present paper (140 sentences) were first translated with Deepl from French to English and then corrected by a professionnal translator. Comparison of the original translation and corrected version reached to 67 BLEU.} and comments in the press point in this direction. 
The company asserts that one of the explanations for its success is the substantial size of the parallel corpus on which the system is trained. 
This corpus, Linguee, contains a billion aligned examples (we assume that these are bi-sentences) \footnote{https://www.linguee.com/, 2020/03/08}. 

These claims naturally have a publicity effect; in reality we don't know precisely what percentage of this corpus is used to translate between two given languages, nor how the corpus is used. 
But this figure is symptomatic of a trend in the field of machine translation: the race for gigantism.
The implicit idea is that increasing the quality of translations requires an increase in the volume of data processed. 
A corollary is that the necessary computing power also increases. 
What is remarkable is that, in both academia and industry, there is a growing interest in increasingly powerful systems. 
\citet{MassivMultilgNMT} works on corpora of about 200 million examples
, \citet{massivMultilgNMTinthewild} on a multilingual corpus of 25 billion parallel sentences. 
With these corpora, the quality of translation can reach to around 35 BLEU when working into English, and 40 when working from English (according to graph 2 in \citep{MassivMultilgNMT}). 

In order to understand the gigantic scope of the corpus being processed, it must be brought down to a human scale. Let's consider a corpus of one billion bi-sentences, that is to say 2 billion sentences. If a human needed to consult so many sentences to become a professional translator, he or she would need about 570 years just to read all of them, at the rate of 16 sentences per minute, 10 hours per day and 365 days per year (see the details of this count in the next section). Yet an individual can produce professional-quality translations by the age of 23. 

The gains (BLEU score and human evaluations) observed with these \enquote{in the wild} systems seem to support the \enquote{gigantesque} strategy. But this strategy has its limits. First of all, it implies a considerable human, material and financial investment. It is therefore reserved for the handful of players who have sufficient financial means. We then run the risk of orienting research in the direction suggested by these few players. This direction may produce good results in the immediate future, but there is no guarantee that the progress will continue in the long term. 

The race for gigantism also raises questions of sustainability. The exploitation of neural systems on an industrial scale owes much to advances in hardware, including the use of graphics cards and other parallel processing devices. These devices are known to be very energy-intensive and their consumption naturally increases with the volume of data to be processed. 
This significant increase in power consumption runs counter to the needs of society, as described by environmental scientists \citep{giec2018}. 

\enquote{Gigantism} is undeniably, though not completely, successful. 
Despite the size of their corpus, Deepl, \citet{MassivMultilgNMT} or others have not yet reached the level of quality provided by a professional translator. 
On the one hand, translation fluidity has improved considerably. 
But reliability is still a problem (see for example the discussion in \citep{nmt_fluency_adequacy_2019}). 
If the very large corpora mentioned above are not sufficient to achieve professional quality, but increasing their size will do the trick, how many bi-sentences must then be added to produce professional quality?

Part of the answer is provided by the studies mentioned above. 
With a maximum of 1 billion bi-sentences, Deepl announces a BLEU score of 33. 
Moreover, with 2 billion bi-sentences, Naaven et al.~\citet{massivMultilgNMTinthewild} gets 35 (English-to-any) and 40 (any-to-en) BLEU
\footnote{These scores are not provided explicitly by the authors. We deduce them from section 3.1 of the paper and from Chart 2. In section 3.1, the authors indicate that they use corpora with a maximum size of 2 billion parallel sentences. In Figure 2, we see that, for the largest corpora, the results are between 35 and 40 BLEU. }. 
The quality would thus improve at best by 8 points for 1 billion additional bi-sentences. Let's assume that a professional-quality translation is around 80 points. Let's then be optimistic and assume that the improvement is linear (which no one can predict, for lack of studies). Given that it is necessary to add 1 billion bi-sentences to gain 8 points, this means that going from 40 to 80 BLEU points would require the addition of 5 billion bi-sentences to the original 2 billion used to reach 40 points. Ultimately, in the best case scenario, 7 billion bi-sentences would be needed to obtain a professional-quality translation. 

This quite considerable figure raises another problem. Since quality bilingual data are not available in sufficient quantity, increasingly degraded-quality data must be incorporated to expand the corpus. When data is collected on the web, there is even the risk of integrating examples that have already been automatically translated by the translation system itself \citep{mosesmtmarketreport2015}. The management of these very large corpora in turn requires the development of pre-processing techniques. And these 
techniques in turn consume energy and resources. 
Accordingly, it is not a virtuous circle. 

These reserves encourage the simultaneous exploration of other avenues. Given the techniques available today, however, any new approach will need a corpus. 
In a frugal context, what is the maximum corpus size to be set, then, and based on which criteria?

\section{Alternative}

The corpus size required varies, of course, depending on the end use of the translation system: translation of specialised or very general texts, multilingualism or not, etc. Our goal is to have an automatic translator that satisfies as many people as possible. We can assume that most end users (individuals, companies) of machine translation need a bilingual translation. They rarely need several languages at the same time. At worst, a multilingual translation can be divided into bilingual translations if the quality of translation from/to pivot language is sufficient \citep{currey-heafield-2019-zero}. Bilingual resources are therefore necessary, first and foremost. 

In addition, we aim for professional quality. This means that the language must be fluent and the content reliable. Defining an intermediate level of quality seems paradoxically more difficult to us, since there are so many possible variations.

To decide on the maximum size of the corpus, we propose to align ourselves with the only \enquote{system} capable of producing a professional-quality bilingual translation today: the human one. The maximum size of the corpus is the number of sentences a professional translator has encountered (through hearing or reading) during his or her entire training. 

Comparisons between humans and machines have some limitations, of course. 
To paraphrase Richard Feynman, making a flying machine does not require feathers or flapping wings. 
But we could counter that using wings that imitate many  <<natural>> flying machines could be useful. 
In other words, the comparison does not necessarily provide solutions, but it could. 
It would be a pity to reject solutions inspired by the observation of nature (indeed, the human professional translator) out of hand. 
Biomimetism has been the source of multiple practical inventions ... including artificial neural networks. 

\section{Frugality concerning corpus: state of the art}

We now evaluate the quantity of corpus that is neededd to become a professional translator, that translate between two languages. 

For convenience, we will use the verb \enquote{acquire} instead of \enquote{encounter}. 
The \enquote{sentences} can be syntactically complete or not. 
We consider that the training begins in the first year of life, since these sentences participate to the linguistic development of the individidual, the hability to process language, and ultimately the hability to translate. 
There are many academic and non-academic paths to become a professional translator, including the cases of bilinguals. Let choose a realistic cursus that can easily be quantified. 
To simplify, let's consider a professional translator who is not natively bilingual and who has acquired a non-mother tongue/ target language during higher education. We consider that this language has not been encountered so far. By following a standard academic training, an individual can start working as a translator at the age of 23, after 5 years of higher education\footnote{These figures correspond to the acquisition of a language like Japanese in French higher education (University).}. 

We do not take into account the learning of foreign languages in middel and high school. 
For example, in France, most children learn English as a second language, and German or Spanish as a third language. 
Taking this learning into account does not make sense, however, since it is impossible to measure its impact on language acquisition at a professional level at university. 
There is no study concerning a correlation between linguistic proximity of the languages acquired, time spent learning, individual skills, the level and quality of teaching, extracurricular practice, etc. 
In addition, there are huge disparities between people, and drawing up a typical, representative profile seems impossible. 
However, in an attempt to eliminate the effect of early foreign language acquisition, we have chosen a scenario in a European context, where the language acquired at university (Japanese) is rarely taught in secondary school. 
In addition, it is linguistically distant from the (Indo-European) languages usually taught in secondary education.

We propose two calculations. First, we calculate the maximum number of bi-sentences that an individual can acquire by the age of 23. 
The results are given in Table \ref{tab1}. 
We first estimate the daily acquisition time, then we deduce the number of sentences acquired annually. 

Our calculation method is very similar to the method used by \citep{compdatarequirements}, who compares the needs of a machine and those of a human in speech recognition. 
Let's consider the number of hours of sleep per day (column 3)\footnote{Figures depend on several socio-cultural and physiological criteria; We have selected the values provided by French pediatricians (http://sommeil.univ-lyon1.fr/ articles/ cfes/ sante/ enfant.php; 2020/10/01); Academic litterature \cite{QuelsommeilaquelageMARTIN2011233} provides similar values.}. 
We deduce the number of waking hours. We consider that, of these waking hours, 2 hours are without acquisition (spent in silence, or taking in sounds other than language). 
The daily acquisition time is provided in col.4. We then calculate the maximum number of sentences that can be acquired during this time (daily aquired number of occurrences of sentences in col.5; annual cumulation is provided in col.6-9). 
We base ourselves on the average reading speed. 
A normal reader can read 250 words in one minute (from \citealp{speedreading}). At an average of 15 words per sentence \citep{mathematicallinguistics}, this corresponds to 16 sentences per minute. 
We deduce that, in 23 years, the young professional translator may have acquired 47.5 million bi-sentences. 
We have to pay attention to the fact that this is the number \enquote{of occurrences} of sentences. 
Since it is impossible to determine the number of repeated sentences, we do not make that distinction. 
In reality there are certainly fewer original sentences. 

Of course, this calculation is based on unrealistic assumptions from a developmental point of view. 
Even a bilingual child does not hear or read \enquote{bi-sentences} but rather monolingual, at best comparable, corpora. This figure is therefore a maximum \enquote{ideal} limit with regard to current techniques in machine translation, which are largely based on bilingual corpora. 

The second calculation is based on more realistic data. Its distinctive feature is that it integrates monolingual corpora. Let's consider that, up to 18 years old, the individual acquires only the corpus of his mother tongue L1. During the five years of higher education, he acquires alternately L1 and a target language L2. Part of the training in higher education consists of learning to match L1 and L2 sentences. Consequently, during his 5 years of university, a quarter of the time is still devoted to the acquisition of L1 (equated to a monolingual corpus), a quarter to the acquisition of L2 (monolingual), and the rest to the acquisition of L1-L2 bi-sentences (bilingual corpus). At the age of 23, when he becomes professional, our individual has: a monolingual L1 corpus of 75.5M sentences, a monolingual L2 corpus of 6M utterances, and a bilingual corpus of 6M parallel sentences. 

\begin{table*}
\centering
\begin{tabular}{|c	|	c	|	c	|	c	|	c	|	c	|	c	|	c	|	c	|}
\hline
&		&			&	awake&	occ.	&	\multicolumn{4}{|c|}{Cumul. per year	(Million sent.)}						\\ \cline{6-9}
&	Age	&	Sleep	&	time&	sent.	&	Analysis 1	&	\multicolumn{3}{|c|}{Analysis 2}				\\ \cline{6-9}
&		&	Hours	&	-2H	&	per day	&	bi-sent.	&	L1	&	L2	&	bi-sent.	\\ \hline
		Infant	&	1	&	16	&	6	&	5,760	&	1.1	&	1.1	&		&		\\ \hline
		Young child	&	2	&	16	&	6	&	5,760	&	2.1	&	3.2	&		&		\\ \hline
			&	3	&	14	&	8	&	7,680	&	3.5	&	6.0	&		&		\\ \hline
			&	4$\sim$5	&	12	&	10	&	9,600	&	5.3	&	9.5	&		&		\\ \hline
		Beg.prim.school	&	6	&	12	&	10	&	9,600	&	8.8	&	16.5	&		&		\\ \hline
			&	7$\sim$9	&	11	&	11	&	10,560	&	10.7	&	20.3	&		&		\\ \hline
		End prim. school	&	10	&	11	&	11	&	10,560	&	16.5	&	31.9	&		&		\\ \hline
			&	11	&	10	&	12	&	11,520	&	18.6	&	36.1	&		&		\\ \hline
			&	12	&	10	&	12	&	11,520	&	20.7	&	40.3	&		&		\\ \hline
			&	13	&	9	&	13	&	12,480	&	23.0	&	44.9	&		&		\\ \hline
		End middle school	&	14$\sim$17	&	8	&	14	&	13,440	&	25.4	&	49.8	&		&		\\ \hline
		End high school	&	18	&	8	&	14	&	13,440	&	35.2	&	69.4	&		&		\\ \hline
			&	19$\sim$20	&	8	&	14	&	13,440	&	37.7	&	70.6	&	1.2	&	1.2	\\ \hline
		End Bachelor	&	21$\sim$22	&	8	&	14	&	13,440	&	42.6	&	73.1	&	3.7	&	3.7	\\ \hline
		End Master	&	23	&	8	&	14	&	13,440	&	47.5	&	75.5	&	6.1	&	6.1	\\ \hline

\end{tabular}
\caption{Cumulative number of sentences read or heard by a professional translator\label{tab1}}
\end{table*}

\section{Discussion and perspectives}

In view of the success achieved in ``real world'' by several machine translation systems, the choice of gigantesque training corpus cannot be condemned. But this success is relative and comes at an exorbitant cost in terms of resources (data and calculation, and consequently hardware).

In this context, many questions arise: 
Is gigantism absolutely necessary to obtain a professional-quality translation?
Can we establish how many examples are theoretically sufficient to train an NMT system capable of providing profesionnal quality translations ?
Which pairs of languages could be concerned?
There are probably many ways to define such a limitation, depending on the purpose of the translation system.

Among many possible solutions, we suggested to take human professional translator as a criteria. 
According to this criterion, a good translation could be accessible with corpora that are significantly smaller than those used by state-of-the-art systems today.
Given the quality resources freely available at present, the first proposal (47.5 million bi-sentences) is very difficult to reach for most language pairs. The second proposal contains mainly monolingual corpora. It is accessible for a larger number of languages. Indeed, large monolingual corpora are available for many languages. Several language pairs have also corpora of at least 6 million bi-sentences. And for several others, it should be easy to build corpora by different techniques (paraphrase, synthetic corpora), with a lesser but still acceptable quality. 

Our purpose here was not to provide a technique to use such <<small>> corpora. 
We can nevertheless suggest a few ideas in relation to the second proposition, where monolingual corpora play an important role.
This suggests that acquiring a strong knowledge of at least the target language should be beneficial. 
Consider the human (professional) translator. S/he usually translates from their foreign language into their native language. The foreign language is the <<source>> language, while the native language is the <<target>> language. 
For example, a native French speaker proficient in Japanese will usually translate from Japanese into French. 
It would then be beneficial to have an excellent knowledge of the French language. 

Indeed, human translators use many extralinguistic data in the course of their work: semantics, encyclopaedic knowledge, contextual knowledge, etc. 
Those data could not be quantified, and must be added to a frugal corpus. 
We can then consider that our comparison underestimates the real quantity of data used by human translators.
Nevertheless, the quantity of resources needed by gigantic corpora is also underestimated. 
Those corpora cannot be saved in a simple file on a simple plug-and-play hard drive. 
Building and maintaining them requires additional specific operations and hardware that are not necessary for <<small>> corpora. 
Such additional data should be taken into account when comparing human translators with machines, even if they do not constitute a corpus, strictly speaking.

We wondered if taking the energy consumption into account would help us to get an idea of the total amount of resources (data like corpora and extralinguistic knowledge, but also calculations) required for training systems. Yet again, a first attempt at calculation shows that the human translator is infinitely more frugal in this respect. Accordingly, using extralinguistic knowledge in addition to corpora does not seem to change the conclusion that a frugal corpus (as defined in this paper) can produce high-quality translations.

\bibliography{arxiv}
\bibliographystyle{acl_natbib}

\end{document}